\documentclass[conference]{IEEEtran}

\usepackage{cite}
\usepackage{amsmath,amssymb}
\usepackage{graphicx}
\usepackage{array}
\usepackage{booktabs} 
\usepackage{url}
\usepackage[T1]{fontenc}
\input{glyphtounicode}
\usepackage[hidelinks,bookmarks=false]{hyperref}
\IEEEoverridecommandlockouts
\usepackage[table]{xcolor}
\usepackage{threeparttable}

\begin{document}

\title{FPGAFly: A 166.7k-Neuron Fruit-Fly Connectome\\
on One FPGA via Deferred Spike Expansion\\
and Shared-Codeword States}

\author{
    \IEEEauthorblockN{
        Ye Ziyang
    }
    \IEEEauthorblockA{
        Department of Electrical Engineering and Information Systems\\
        Graduate School of Engineering, The University of Tokyo\\
        Tokyo, Japan\\
        \href{mailto:ye_zy@silicon.t.u-tokyo.ac.jp}{ye\_zy@silicon.t.u-tokyo.ac.jp}
    }
    \thanks{This work has been submitted to IEEE for possible publication. Copyright may be transferred without notice, after which this version may no longer be accessible.}
}

\maketitle

\begin{abstract}
Synapse-resolution fruit-fly (Drosophila) connectomes enable neuron-by-neuron simulation of entire insect nervous systems, yet whole-brain models still require computationally demanding software or multiple neuromorphic chips, and single FPGAs support only small or statistical networks. Scaling to the full connectome, per-synapse spike expansion scales queue traffic with fan-out up to 11.2k, while 41-bit neuron states require 6.83~Mbit, exceeding the 6.27~Mbit on-chip block static RAM (BSRAM) of the target device. FPGAFly runs the 166.7k-neuron, 25.6M-connection leaky integrate-and-fire male central nervous system (CNS) connectome on one FPGA. Order-preserving deferred expansion queues one 64-bit descriptor per spike in source-ID order, cutting external memory traffic by 54.5\%--57.9\% and peak queue data by 95.6\%--96.5\% with bit-identical results. A 24-bit encoding shares codewords across normal, refractory, and quantized out-of-range states, reducing BSRAM usage from 438 to 270 with 1245 extra LUTs and zero changes across 10,000 MNIST predictions. At 80~MHz on a Gowin GW5AST-LV138, each 100~ms window finishes in 86.83~ms on average and within 93.05~ms across 2000 tests, realizing the first complete Drosophila CNS connectome simulation on one FPGA.
\end{abstract}

\begin{IEEEkeywords}
Drosophila connectome, large-scale neural simulation, memory-efficient architecture, neuromorphic computing, spiking neural networks  
\end{IEEEkeywords}

\section{Introduction}

Synapse-resolution connectomes now cover the complete adult Drosophila brain~\cite{dorkenwald2024} and the entire male central nervous system (CNS), including the ventral nerve cord~\cite{berg2026}. Connectome-derived networks yield experimentally confirmed predictions of taste-driven feeding~\cite{shiu2024} and visual activity~\cite{lappalainen2024}, while neuromechanical models simulate walking and flight under multisensory input~\cite{lobatorios2022,wangchen2024,vaxenburg2025}. Coupling these models with real-world sensors and actuators would require real-time execution.

\begin{figure*}[!t]
\centering
\includegraphics[width=0.9\textwidth]{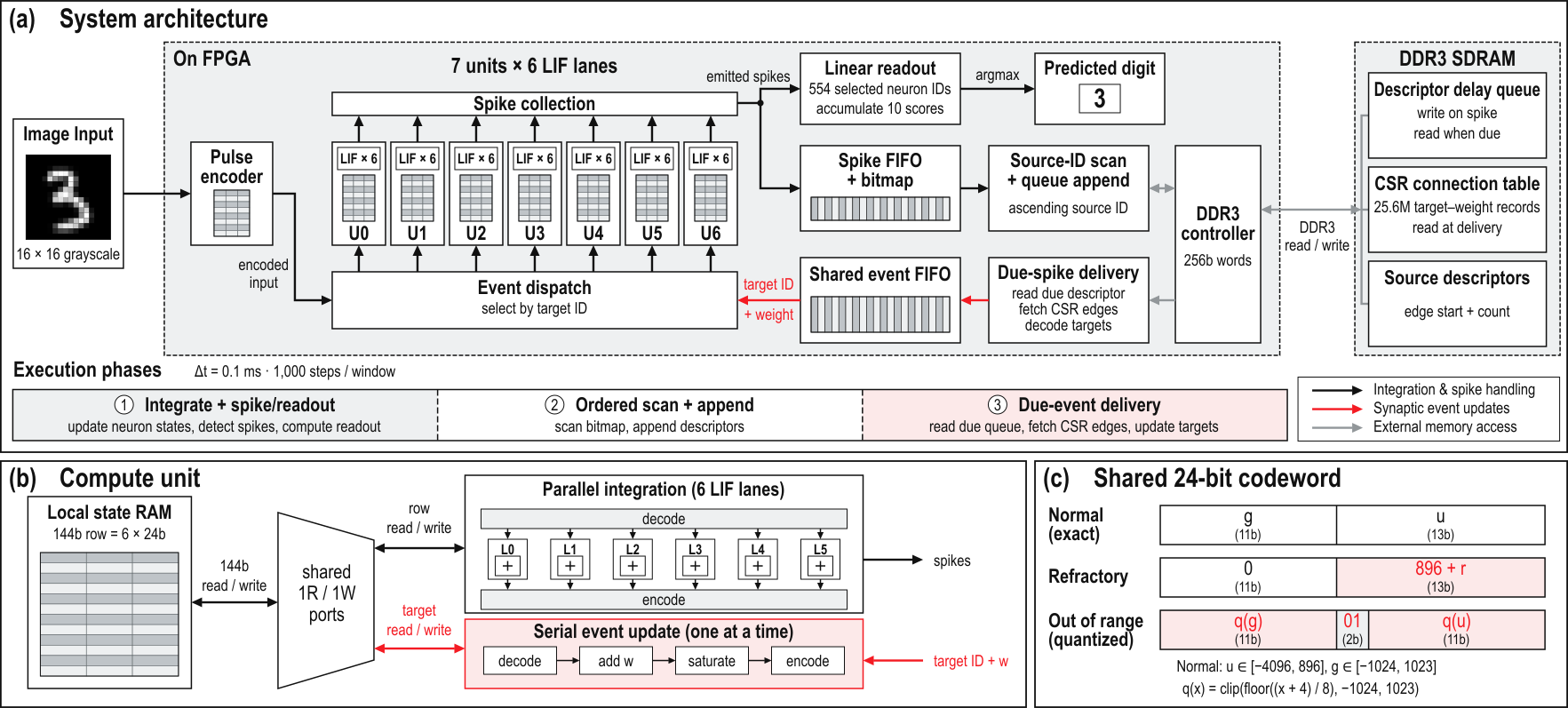}
\caption{FPGAFly architecture. (a) System dataflow and execution phases. (b) Compute unit with parallel integration and serial event updates. (c) Shared 24-bit state encoding.}
\label{fig:arch}
\end{figure*}

Large-scale simulators support statistical networks~\cite{rhodes2020realtime,kauth2023}. Pattern-based and procedural connectivity reduce storage~\cite{wang2018cortex,knight2021}, whereas connectomes require explicit per-connection storage. Software simulations of whole-brain Drosophila models still run 4.4--14$\times$ slower than real time in Brian~2~\cite{shiu2024,stimberg2019,wang2025loihi}. While neuromorphic chips~\cite{merolla2014,davies2018,furber2014,benjamin2014} and multi-FPGA systems~\cite{moore2012,cheung2016} provide high capacity, simulating the 140k-neuron FlyWire brain requires 12 Loihi~2 chips~\cite{wang2025loihi}, and single FPGAs support only trained networks~\cite{neil2014,li2024} or a 2.2k-neuron olfactory circuit~\cite{nanami2024}.

Memory is the primary bottleneck. Expanding each spike at emission into per-synapse delay-queue records scales external memory traffic with fan-out, which is heavy-tailed in fly connectomes~\cite{lin2024} and reaches 11.2k here. Meanwhile, storing two 18-bit neuron states and a 5-bit refractory counter requires 41 bits per neuron, or 6.83~Mbit for the 166.7k-neuron CNS, exceeding our target FPGA's 6.27~Mbit of block static RAM (BSRAM). Off-chip storage is impractical because updating every neuron each 0.1~ms step requires about 17~GB/s of state traffic, 6.7$\times$ the DDR3 controller's 2.56~GB/s peak.

Layerwise dataflows~\cite{li2024,narayanan2020} require a feed-forward order that recurrent connectomes lack. Address-event communication~\cite{boahen2000}, source-indexed row lookup~\cite{rhodes2018spynnaker}, and hierarchical routing memories~\cite{moradi2018} avoid per-synapse queue records, but our model saturates and quantizes states on each delivered event, so queueing spikes in parallel lane completion order alters trajectories. For state storage, narrowing independent fixed-point fields~\cite{hopkins2020} either fails 80~MHz timing at 29 bits or distorts dynamics at 24 bits.

FPGAFly\footnote{Source code: \url{https://github.com/ZiyangYE/FPGAFLY_ISCAS2027}} exploits model delay and reset bounds to fit on a single Gowin GW5AST-LV138 FPGA, as shown in Fig.~\ref{fig:arch}. Uniform 18-step source delays and 22-step refractoriness allow bitmap-ordered deferred expansion to queue at most one 64-bit descriptor per neuron, cutting external memory traffic by 54.5\%--57.9\% with bit-identical results. Firing and refractory resets leave unused codes for a 24-bit encoding to pack refractory and quantized out-of-range values alongside normal states, reducing BSRAMs from 438 to 270 with 1245 extra LUTs and zero changed MNIST predictions. At 80~MHz, FPGAFly finishes 2000 100~ms windows in 86.83~ms on average and within 93.05~ms, realizing the first complete Drosophila CNS connectome simulation on one FPGA.

\section{Model and Architecture}

FPGAFly time-multiplexes all neurons across seven 6-lane units with local BSRAM state memory (Fig.~\ref{fig:arch}(a)), while external DDR3 memory holds the connection table, source descriptors, and delay queue. Each 0.1~ms step executes neuron integration and readout, ordered source-ID scanning and queueing, and due-event delivery. At delivery, due descriptors expand into target-weight events through a shared FIFO and event dispatcher to update neuron states.

\subsubsection{Neuron Model}

The network is a frozen fixed-point leaky integrate-and-fire model derived from MaleCNS v1.0~\cite{berg2026}. A compressed sparse row (CSR) table stores its 25.6M neuron-pair connections with signed 18-bit weights. The membrane state $u$ and synaptic state $g$ carry seven and six fractional bits, respectively, and $r$ counts refractory steps. With the rounding \mbox{$R(x)=\lfloor (x+2^{17})/2^{18}\rfloor$}, a nonrefractory neuron is updated as
\begin{equation}
u'=u+R(1307(I-u)+2588g),\;\; g'=g+R(-5191g),
\end{equation}
where $I$ is the input drive and the constants are exponential integration factors for $\tau_m=20$~ms and $\tau_g=5$~ms at $\Delta t=0.1$~ms, scaled by $2^{18}$. The factor $2588$ absorbs the 1-bit alignment between $g$ and $u$. If $u'>896$, the neuron spikes, resets $u=g=0$, and remains refractory for 22 steps. Otherwise, both states saturate to 18 bits before state encoding. All synapses share an 18-step (1.8~ms) delay. Each due event adds its weight $w$ to a nonrefractory target's $g$, followed by saturation, encoding, and immediate writeback. For stimulus input, lamina neurons receive constant background drive $I$, and a pulse encoder maps $16\times16$ grayscale images to 3335 visual input neurons.

\begin{table*}[!t]
\centering
\caption{Resource Usage, Timing, and Fidelity of State-Format and Spike-Expansion Variants}
\label{tab:variants}
\footnotesize
\setlength{\tabcolsep}{3pt}
\renewcommand{\arraystretch}{1.18}
\begin{threeparttable}
\begin{tabular}{@{}l ccc c ccccc@{}}
\toprule
& \multicolumn{3}{c}{Synthesis}
& P\&R at 80~MHz
& \multicolumn{5}{c}{Fidelity vs.\ uncompressed format}\\
\cmidrule(lr){2-4}\cmidrule(lr){5-5}\cmidrule(lr){6-10}
Variant
& BSRAMs & LUTs & Registers
& Setup WNS (ns)
& Accuracy
& \shortstack{Changed\\predictions}
& \shortstack{Rate MAE\\(Hz)}
& \shortstack{Population\\correlation}
& \shortstack{Spike\\match}\\
\midrule
Uncompressed (41-bit)
& 438 & 37,953 & 56,328
& \textit{exceeds capacity}
& 86.91\% & \multicolumn{4}{c}{\textit{reference}}\\

\rowcolor{black!6}
\textbf{Shared 24-bit (proposed)}
& \textbf{270} & 39,198 & 48,272 & \textbf{+0.020}
& 86.91\% & 0 & 0.0709--0.231
& 0.9904--0.9995 & 42.17\%\\

Shared 24-bit, early expansion
& 270 & 39,859 & 48,127 & +0.038
& \multicolumn{5}{c}{\textit{bit-identical to shared 24-bit on six inputs}}\\

Independent 29-bit
& 326 & 35,236 & 48,805 & $-0.653$
& 86.92\% & 1 & 0.491--1.30
& 0.9868--0.9994 & 42.24\%\\

Independent 24-bit
& 270 & 34,688 & 45,736 & \textit{route failed}
& 87.82\% & 987 & 9.64--18.5
& $-0.0017$--0.4053 & 1.06\%\\
\bottomrule
\end{tabular}

\par\smallskip\noindent
Synthesis settings are identical, and expansion is deferred unless noted.
WNS, worst setup slack after place-and-route.
\end{threeparttable}
\end{table*}

\subsubsection{Order-Preserving Deferred Expansion}

When a source spikes, early expansion queues one target-weight record per outgoing synapse, whereas deferred expansion queues a single 64-bit descriptor (Fig.~\ref{fig:queue}(a)). Each spiking source's 64-bit entry is fetched from the DDR3 source-descriptor table and queued directly, packing a 25-bit CSR edge index, 14-bit out-degree, 14-bit coefficient start, and 4-bit coefficient count. Both schemes pack four 64-bit records per 256-bit memory word.

Per-event saturation and quantization make updates order-dependent. Across the seven 6-lane units, any 42-neuron row with a spike pushes a 12-bit row index and 42-bit mask into a 64-entry FIFO. Draining this FIFO updates readout scores and marks source IDs in a bitmap. Scanning this bitmap in ascending order then enqueues descriptors deterministically, regardless of lane completion order. Since all outgoing synapses of a source share one delay, both schemes deliver the exact same event sequence and keep states bit-identical at every step.

Because the 22-step refractory period exceeds the 18-step delay, each neuron has at most one descriptor in flight, bounding valid queue data to 1.27~MiB versus 195.2~MiB for early expansion.

\subsubsection{Shared-Codeword State Encoding}

Refractory neurons hold $u=g=0$, allowing state encoding without a separate refractory field (Fig.~\ref{fig:arch}(c)) as
\begin{equation}
E(u,g,r)=
\begin{cases}
0_{11}\Vert[896+r]_{13},&1\le r\le22,\\
[g]_{11}\Vert[u]_{13},&r=0,\ (u,g)\in\mathcal N,\\
[q(g)]_{11}\Vert01\Vert[q(u)]_{11},&\text{otherwise},
\end{cases}
\end{equation}
where $\mathcal N$ denotes $u\in[-4096,896]$ and $g\in[-1024,1023]$, $q(x)=\operatorname{clip}(\lfloor (x+4)/8\rfloor,-1024,1023)$, $[x]_k$ is a $k$-bit two's-complement field, and $\Vert$ denotes concatenation. Spiking caps nonrefractory $u$ at 896, so the lower 13-bit field uses 897--918 to encode $r$ and 2048--4095 to mark out-of-range states quantized by 8.

Each unit stores six 24-bit states per 144-bit row across sixteen 9-bit banks. Integration and event updates use state RAM in separate phases
(Fig.~\ref{fig:arch}(b)). Each unit integrates six neurons in parallel
and updates target neurons serially in 4--5 cycles per event. Domain $\mathcal N$ and refractory codes span $4993\times2048+22\approx 2^{23.3}$ states per neuron. Two depth-cascaded $2048\times9$-bit BSRAMs per bank provide 4096 rows, totaling 224 state BSRAMs across seven units plus 46 for queues, bitmap, and memory interfaces. Each event writes to only the three or four banks that span the target lane.

\section{Evaluation Methods}

We implemented FPGAFly in Gowin EDA V1.9.12.03 on a Tang Mega 138K Pro board with a Gowin GW5AST-LV138 FPGA containing 340 BSRAMs. Core and DDR3 clocks run at 80 and 320~MHz, giving 8000 core cycles per 0.1~ms step. Each $16\times16$ area-averaged MNIST image is presented over a 100~ms test window of 1000~steps. States and queues persist within continuous runs. For classification, a linear readout trained offline on 50k MNIST images accumulates ten class scores across 554 selected neurons.

\begin{figure}[!t]
\centering
\includegraphics[width=0.95\columnwidth]{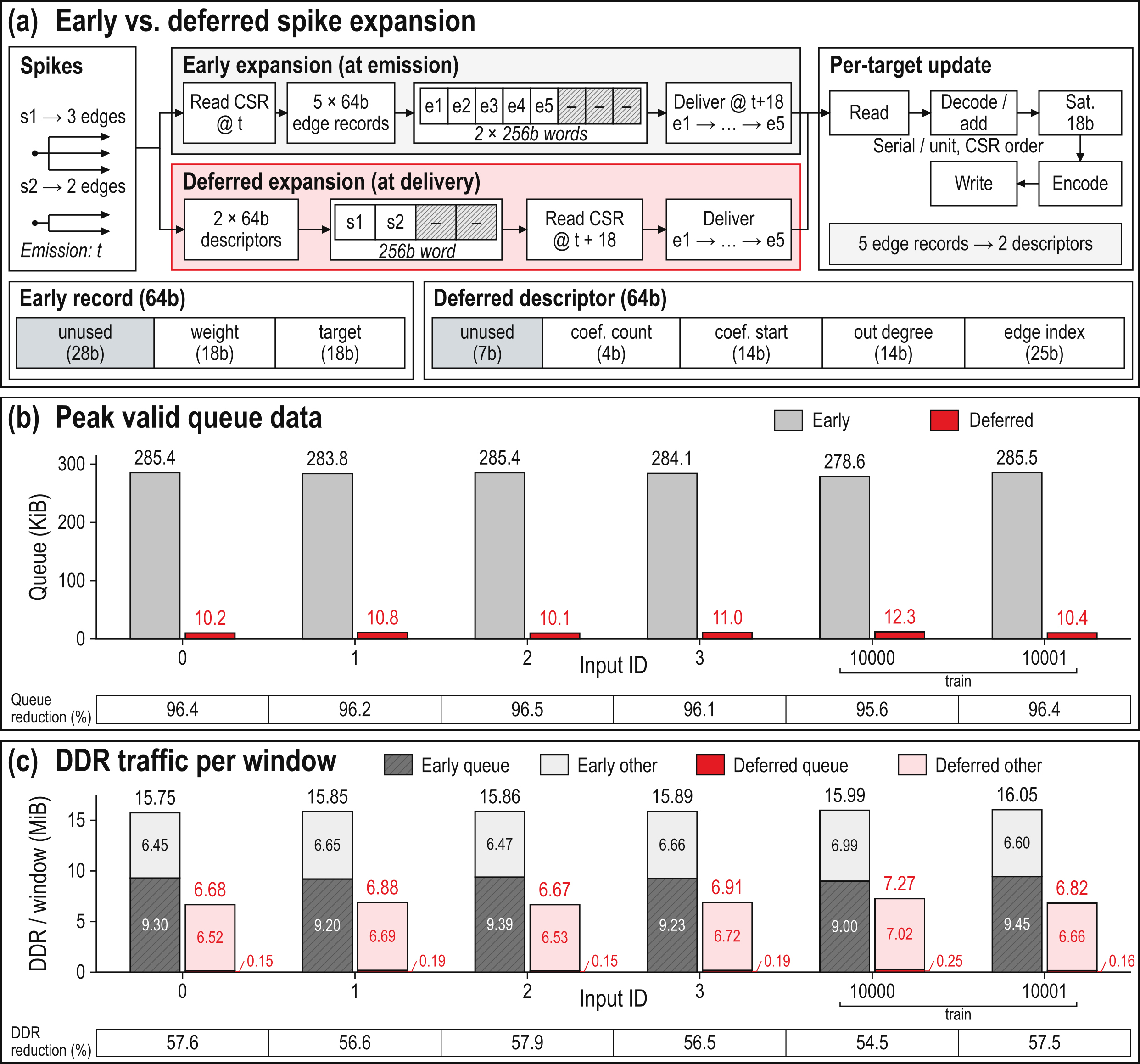}
\caption{Early versus deferred spike expansion. (a) Record formats and dataflow. (b) Peak valid queue data and (c) DDR traffic per window.}
\label{fig:queue}
\end{figure}

\begin{figure*}[!t]
\centering
\includegraphics[width=0.9\textwidth]{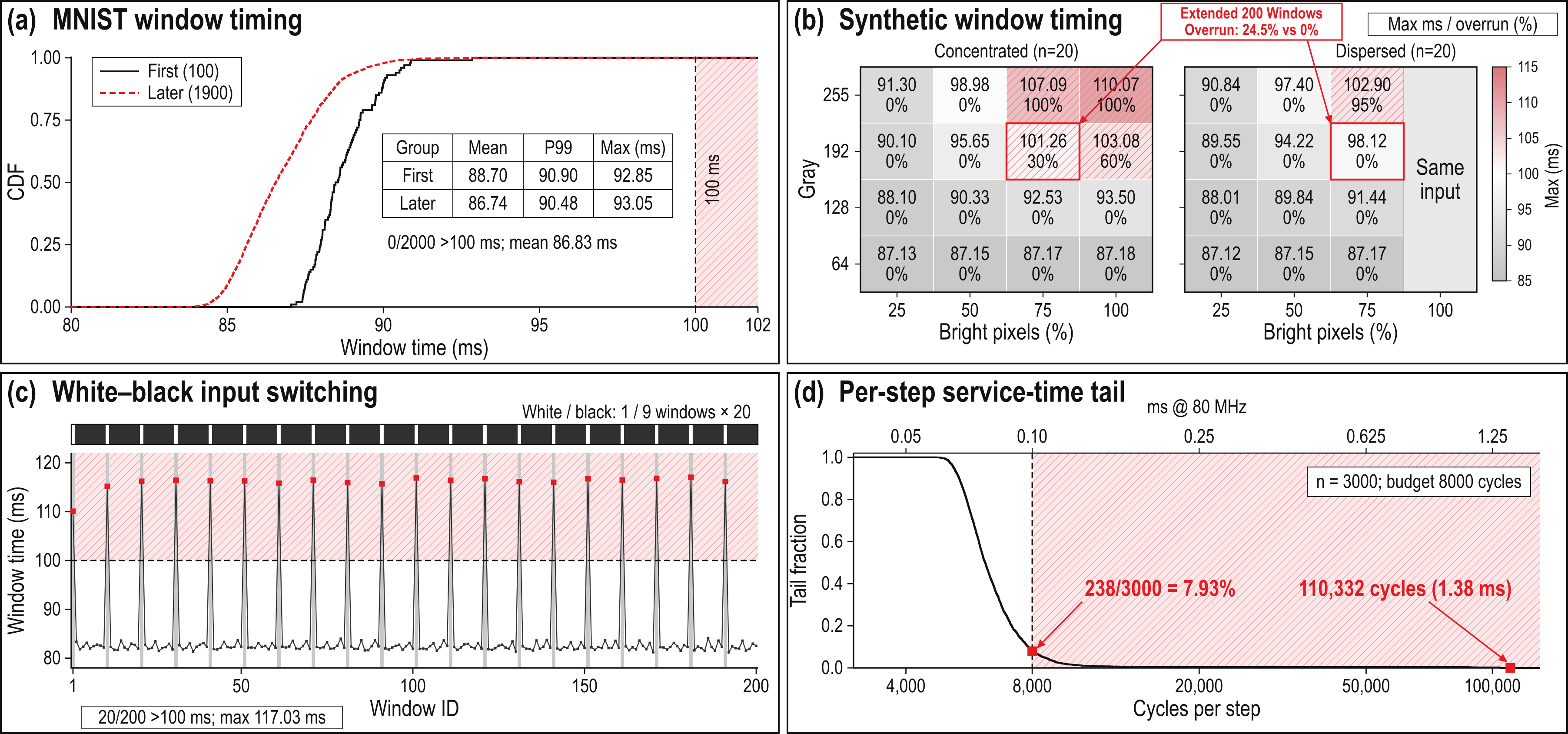}
\caption{Board window timing and simulated per-step service time. (a) Window time over 2000 MNIST windows. (b) Maximum window time and overrun rate for synthetic inputs. (c) Window time with one white per nine black windows. (d) Per-step service time in RTL simulation over 3000 steps. }
\label{fig:board}
\end{figure*}

\begin{table*}[!t]
\centering
\caption{Comparison With Prior Large-Scale Spiking-Network and Connectome Simulation Platforms}
\label{tab:comparison}
\footnotesize
\setlength{\tabcolsep}{2.5pt}
\renewcommand{\arraystretch}{1.1}
\begin{threeparttable}
\begin{tabular}{l l c l c c c c}
\toprule
Work & Platform & Devices & Network & Neurons & Connections & Time step & Real-time factor\\
\midrule
\multicolumn{8}{l}{\textit{Statistical or procedural networks}}\\
Rhodes \textit{et al.}~\cite{rhodes2020realtime} & SpiNNaker & Multi & Cortical microcircuit & 77k & 0.3B & 0.1~ms & 1.0\\
Kauth \textit{et al.}~\cite{kauth2023} & Xilinx Virtex-7 XC7VX690T & 35 & Cortical microcircuit & 77k & 0.3B & --- & 0.05\\
R. Wang \textit{et al.}~\cite{wang2018cortex} & Intel Stratix V 5SGXEA7 & 1 & Cortex (pattern-based) & 20M--2.6B & --- & 1~ms & 1.0\\
Knight and Nowotny~\cite{knight2021} & NVIDIA TITAN RTX & 1 & Macaque cortex (procedural) & 4.13M & 24.2B & --- & ---\\
\midrule
\multicolumn{8}{l}{\textit{Measured connectomes}}\\
Brian 2~\cite{shiu2024,stimberg2019,wang2025loihi} & CPU & --- & FlyWire brain & $\sim$140k & $\sim$15M & 0.1~ms & 4.4--14\\
F. Wang \textit{et al.}~\cite{wang2025loihi} & Intel Loihi 2 & 12 & FlyWire brain & $\sim$140k & $\sim$15M & 0.1~ms & 0.054--5.8\\
Nanami \textit{et al.}~\cite{nanami2024} & Xilinx Artix-7 XC7A35T & 1 & Olfactory circuit & $\sim$2.2k & --- & 1~ms & 1.0\\
\rowcolor{black!6}
\textbf{This work} & \textbf{Gowin GW5AST-LV138} & \textbf{1} & \textbf{MaleCNS brain + nerve cord} & \textbf{166.7k} & \textbf{25.6M} & \textbf{0.1~ms} & \textbf{0.87 / 0.93}\\
\bottomrule
\end{tabular}
\par\smallskip\noindent
Real-time factor is wall-clock time per unit of simulated time. Brian 2 and Loihi 2 values from~\cite{wang2025loihi} are activity-dependent. This work reports mean / maximum over 2000 MNIST windows.
\end{threeparttable}
\end{table*}

\subsubsection{Queue Organization} Early and deferred expansion shared the encoding, bitmap scanner, and datapath. RTL simulation compared both schemes on test inputs 0--3 and training inputs 10000--10001. Valid queue data comprised all in-flight records, while DDR traffic measured 32~B per accepted command across all phases.

\subsubsection{State Formats} The 41-bit uncompressed baseline uses 18-bit $u$ and $g$ states matching the DSP input width plus a 5-bit refractory counter. An independent 29-bit format with 13-bit $u$ and 11-bit $g$ preserves fractional precision but clips outside its narrower range, while an independent 24-bit format with 10-bit $u$ and 9-bit $g$ was optimized for accuracy on 200 training images. Software simulations evaluated single 100~ms windows across all 10,000 test images and 20~s streams of all-black, normal MNIST, all-white, and alternating white-black inputs. Across the four streams, we report the min--max ranges for both the firing-rate mean absolute error (MAE) over active neurons and the population correlation in 5~ms bins. Classification accuracy used all 10,000 test windows. Exact spike match used the 20~s MNIST stream.

\subsubsection{Implementation and Board Timing} Ten test images per class each ran for 20 consecutive windows, totaling 2000 windows. Stress tests used 60~s of all-white input and 20~s of alternating windows with a 1:9 white-to-black ratio. We also evaluated 28 synthetic patterns in which 25\%--100\% of the pixels are concentrated or dispersed with gray levels from 64 to 255, while the remaining pixels are set to zero. Each synthetic pattern ran for 20 windows, extended to 200 windows for the 75\% gray-192 pair. RTL simulation tracked per-step service time over three windows, or 3000 steps, of MNIST test image 0. Window time ran from initialization, or from the first step on later windows, until classification completed, excluding host I/O. The real-time factor is its ratio to the simulated 100~ms.

\section{Results}

\subsubsection{Queue Organization}

Early and deferred expansion produced bit-identical final states, activity, and scores on tested inputs. Deferred expansion reduced peak valid queue data by 95.6\%--96.5\% and DDR traffic per window by 54.5\%--57.9\% (Fig.~\ref{fig:queue}). Reduced queue traffic drove these savings, while non-queue traffic increased by only 24.5–74.7~KiB over early expansion.

\subsubsection{State Formats}

Only the shared 24-bit format fit on chip, met timing, and preserved all 10,000 predictions, saving 168 BSRAMs at the cost of 1245 extra LUTs (Table~\ref{tab:variants}). Both the shared 24-bit and independent 29-bit formats compute exact in-domain updates. However, out-of-range clipping and quantization cause the state trajectories to diverge~\cite{brette2007}, which reduces the exact spike match to $\sim$42\%. Even so, extending the range through quantization preserves population dynamics and lowers rate MAE by 5.6--6.9$\times$ versus clipping. The independent 24-bit format failed routing, suppressed 62.6\% of spikes, and altered 987 predictions despite accuracy tuning, severely distorting network dynamics.


\subsubsection{Implementation and Board Timing}

After place-and-route, the design met timing at 80~MHz with +0.020~ns setup and +0.131~ns hold slack, using 39.2k LUTs, 49.1k registers, 270 BSRAMs, and 136 DSPs. All 2000 MNIST windows met the 100~ms budget, averaging 86.83~ms and peaking at 93.05~ms (Fig.~\ref{fig:board}(a)). Every window's spike counts, event counts, and scores matched the software model. RTL cycle counts predicted board timing with a median error of 0.082\%.

Under synthetic stress tests, 5 of 28 conditions exceeded the 100~ms budget (Fig.~\ref{fig:board}(b)). At gray level 192 with 75\% bright pixels, concentrated input overran in 49 of 200 windows while dispersed input never overran. Sustained all-white input averaged 108.12~ms, and periodic white-window bursts peaked at 117.03~ms before recovering immediately on subsequent black windows (Fig.~\ref{fig:board}(c)). For MNIST test image~0, 7.93\% of steps exceeded 8000 cycles (Fig.~\ref{fig:board}(d)).


\section{Discussion and Conclusion}

Bitmap-ordered deferral brings deterministic delivery order to source-addressed schemes~\cite{boahen2000,rhodes2018spynnaker} for order-sensitive updates, while refractoriness bounds valid queue data to 1.27~MiB for any input. Shared 24-bit codewords preserve in-domain precision and quantize out-of-range states, saving 168 BSRAMs at a 3.28\% LUT cost while keeping 0.9904--0.9995 population correlation and 0.0709--0.231~Hz rate error. Independent fields either clip out-of-range values at 29 bits using 326 BSRAMs or coarsen in-domain resolution at 24 bits and diverge~\cite{brette2007}. Unlike designs using procedural connectivity~\cite{wang2018cortex,knight2021} or 12 Loihi~2 chips~\cite{wang2025loihi}, FPGAFly is the only single-device platform in Table~\ref{tab:comparison} that executes an entire CNS connectome, supporting a larger graph at the same 0.1~ms step size.

In conclusion, order-preserving deferred expansion and shared-codeword encoding enable the complete 166.7k-neuron, 25.6M-connection Drosophila CNS connectome to run in real time on one low-cost FPGA. Both techniques apply beyond Drosophila to event-driven simulators with per-source delays and neuron models with unreachable post-reset states. Future work will explore motor readout pathways and tighter 0.1~ms per-step tail latency under bursty activity to support closed-loop sensor and robot actuation.


\end{document}